\documentclass[letterpaper, 10pt, conference]{ieeeconf} 

\IEEEoverridecommandlockouts                              

\usepackage{mathptmx} 

\usepackage{mathpazo}                 
\usepackage{helvet}
\usepackage{amsmath}
\usepackage{amssymb}
\usepackage{amsopn}
\usepackage{wrapfig}
\usepackage{rotating}
\usepackage{subfigure}
\usepackage{graphicx}
\usepackage{color}
\usepackage{multirow}
\usepackage{comment}
\usepackage[ruled,vlined]{algorithm2e}
\usepackage{url}
\usepackage{cite}
\usepackage{xcolor}
\usepackage{multirow}
\usepackage[normalem]{ulem}
\usepackage{listings}
\usepackage[colorlinks=false,pdfborder={0 1 1},letterpaper=true,pagebackref=true]{hyperref}
\usepackage{indentfirst}
\usepackage{grffile}
\usepackage[utf8]{inputenc}
\usepackage{breqn}

\newcommand{\OUT}[1]{}
\newcommand{\excise}[1]{}
\newcommand{\prm}{{\sc prm}}

\newcommand{\todo}[1]{\textcolor{red}{To do: #1}}
\newcommand{\jml}[1]{#1}

\newcommand{\out}[1]{}
\newcommand{\revised}[1]{#1}
\usepackage{soul}

\newtheorem{thm*}{Theorem}[section]

\title{\LARGE \bf 
Persistent Covering of a Graph \\under Latency and Energy Constraints}

\author{Jyh-Ming Lien, Sam Rodriguez, and Marco Morales}

\begin{document}
\maketitle


\begin{abstract}
Most consumer-level low-cost unmanned aerial vehicles (UAVs) have limited battery power and long charging time. 
Due to these energy constraints, they cannot accomplish many practical tasks, such as monitoring a sport or political event for hours. The problem of providing the service to cover an area for \revised{an extended} 
time is known as persistent covering in the literature. 
In the past, researchers have proposed various hardware platforms, such as  battery-swapping mechanisms,  to provide persistent covering.
However, algorithmic approaches \revised{are limited mostly} due to the computational complexity and intractability of the problem. Approximation algorithms have been considered to segment a large area into smaller cells that require periodic visits under the latency constraints. However, these methods assume unlimited energy. 
In this paper, we explore geometric and topological properties that allow us to significantly reduce the size of  the optimization problem.
Consequently, the proposed method can efficiently determine the minimum number of UAVs needed  and schedule their routes to cover an area persistently. 
We demonstrated experimentally that the proposed algorithm has better performance than the baseline methods. 
\end{abstract}

\section{Introduction}                                  


Teams of Unmanned Aerial Vehicles (UAVs) have many potential applications such as monitoring and responding to wildfires, crops inspection, or monitoring of crouds. 
A crucial challenge to fully realize such systems is that they operate autonomously and persistently over long periods of time. 
The limited flight endurance of off-the-shelf UAVs, typically between 15 and 30 minutes, and long replenishing time (usually twice \revised{as long as} the flight time) strongly restricts the class of missions that they can carry out. 


\begin{figure}[t]
\centering
\includegraphics[width=0.5\textwidth]{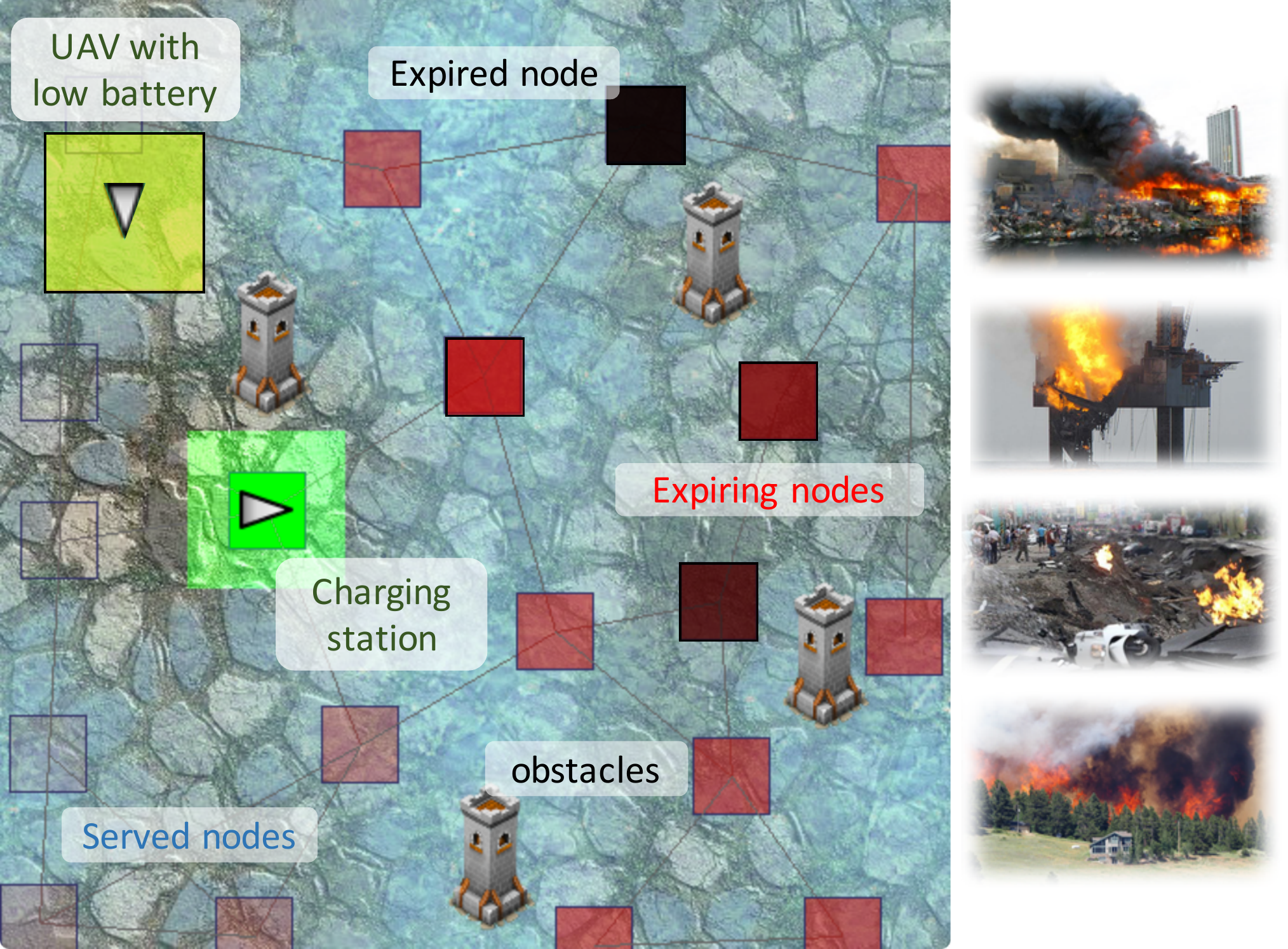}
\caption{How many UAVs with limited battery can cover these regions of interest persistently? \revised{Here}, each UAV must return to the charging station before its battery is depleted and, \revised{as in most of the off-the-shelf quadrotors, charge for a long} time before it can serve again. Each region (squares) must be serviced periodically before its timer expires (obeying the latency constraints). \revised{For example, in a fire emergency, these regions need to be monitored for much longer time than the UAV flying time.}}
\label{fig:uas}
\end{figure}

This paper addresses the questions of finding the minimum number of UAVs that can perform persistent covering on a given graph $G$ 
as in Fig.~\ref{fig:uas}.
While persistent services provided by robots have been studied, 
this paper presents the first work that considers flight endurance, charging time and latency constraints in a unified framework.
Under this framework, we present  a suite of algorithms that pre-process $G$ in order to identify feasible routes (defined as {\em closed walks} in Section~\ref{sec:method}). 
Examples of these routes include segments of loops obtained via the Traveling Salesperson Problem (TSP), \revised{which is a classic graph search problem that seeks to visit all nodes exactly once with the lowest cost},   and  via \revised{a new tour that we call the} {\em maximum lollipop tour} \revised{(due to its shape) as detailed in Section~\ref{sec:ourmethod}. Examples of  maximum lollipop tours can be found in Fig.~\ref{fig:lollipops}}.
\revised{The proposed framework avoids the computation bottleneck faced by previous methods that 
heavily relied on an Integer  Programming (IP) solver to perform the heavy lifting.}
We show that, through these feasible routes, our method finds solutions from seconds to a few minutes instead of hours,  even for graphs with hundreds of nodes.



\section{Related Work}
\revised{The challenges faced to provide robotic persistent services} have been extensively studied. 
When a task (e.g. reaching a goal, coverage, surveillance, or visiting regions at specific times) requires persistency, 
the issue of energy management becomes \revised{central}. This issue \revised{has been}
addressed in both the hardware layer (e.g., charging stations
\cite{kannan2013autonomous} or
swapping devices
\cite{ure2015automated}) and
the software layer (e.g., to determine charging station placement
\cite{yu2019coverage})
for Unmanned Aerial Vehicles (UAVs)
\cite{7101619},
ground mobile robots
\cite{7061469},
aquatic robots \cite{wawerla2007nearoptimal}, or cabled
robotic systems \cite{borgstrom2008energy}.

\jml{This section briefly reviews algorithmic approaches \revised{for} persistent covering and surveillance.
We categorize them into \emph{behavioral methods} that can handle complex scenarios and \emph{optimization-based
methods} that provide theoretical guarantees.}



\subsection{Coverage \& Energy-Aware Coverage}

Coverage, this is, visiting all the regions of interest in an
environment, was initially approached mostly as a geometric
problem such as with the boustrophedon decomposition
coverage \cite{choset2000coverage}. \revised{The coverage
techniques described below} account for energy constraints.

\subsubsection{Behavioral Methods}
\revised{Several coverage algorithms incorporate the energy
constraint although without guarantees on optimality}.
A method that uses a divide-and-conquer
decentralized strategy \cite{mei2006deployment} aims to maximize
sensing while minimizing time and energy.
Another method combines sensor-based reactive behaviors and dynamic feedback
behaviors to produce smooth coverage paths \cite{lee2011smooth}.
One more approach \revised{encodes problem constraints,
available energy, speed and weight through an
energy function~\cite{7101619}}. Another approach
\cite{strimel2014coverage} incorporates available energy into the
edges of an adjacency graph and poses the
problem as the Traveling Salesperson Problem (TSP).

\subsubsection{Optimization-based Methods}
Some coverage methods \revised{that use
optimization functions} provide cost guarantees. In~\cite{kwok2008deployment},
an optimization function that \revised{reflects} the global energy
available to the group  guides motions of
robots that are assigned time-varying regions.
Another approach \cite{sipahioglu2010energy} for the multi-robot case
poses \revised{coverage} as the capacitated arc routing
problem \revised{to find} the shortest tours that \revised{visit} some
arcs. Approaches that provide bounds on path \revised{cost} include
~\cite{shnaps2016online} which maps unknown environments into a cell
decomposition as it is being covered, or ~\cite{wei2018coverage} which partitions the environment into
countour-connected parts where rectiliniar motion is applied. 

Encoding sites and battery levels in graphs is also common, such as
\cite{yu2019algorithms} that poses the problem as the Generalized
Traveling Salesperson Problem (GTSP) to find an optimal tour for a UAV
to visit a set of sites in minimum time with support of UGVs that
recharge them at rendezvous sites. A similar approach
\cite{yu2019coverage} incorporates a Boustrophedon cell
decomposition. A related problem for multiple UAVs to collect maximum
rewards under battery constraints has been posed as the $k$-{\sc
  Stroll} problem (NP-hard) \cite{sadeghi2019minimum}.

\subsection{Surveillance \& Persistent Surveillance}

Surveillance is similar to coverage, but with timing constraints to
ensure that regions are visited periodically. In persistent
surveillance, the time between visits is either bounded or
minimized. Another \revised{frequent goal is to minimize} the maximum time between
visits, also referred as age or latency. 


\subsubsection{Behavioral Methods} 
 
Some approaches seek \revised{to optimize} usage of charging stations through
behaviors based on \revised{the available energy}. In one approach \cite{austin2001mobile}
a fixed energy threshold is set for a robot to recharge, while another
approach \cite{sempe2002autonomous} also considers available stations. Heuristics inspired in animal behaviors \revised{have been}
combined with \revised{thresholds in~\cite{wawerla2007nearoptimal} based} on whether the robot can reach a goal
and a charging station. Another approach \cite{derenick2011energy} is
to weigh a trade-off between achieving coverage and \revised{maintaining} a desired
energy reserve. Work subdivision has been applied with helper robots ready to
replace active robots \cite{strimel2014coverage} and extended with
heuristics \cite{mishra2016battery}.

\subsubsection{Optimization-based Methods}

An approach to provide nearly continuous UAV escort formulates the
problem as a Mixed Integer Linear Program (MILP)\cite{Song2016} to
minimize energy consumption and, at the cost of optimality,
incorporates uncertainties through heuristics. Similarly, a
resolution complete algorithm \cite{peters2019unmanned} poses the problem as
a constrained optimization to minimize mission time and time to reach
a goal and
approximates a discrete search on a graph whose nodes pair
configurations with feasible trajectories.

Among the approaches that aim to minimize latency, we find one that
formulates the problem as two linear programs (one for the speed controller and
another to minimize latency that considers whether
regions are or not in range) \cite{smith2011persistent}. Another
method poses this problem as a TSP where potential recharging points are nodes of a
directed acyclic graph \cite{mathew2013graphbased}. In both cases the
trajectories of servicing UAVs \revised{predetermined}.

In a more realistic setting, the UAV trajectory should be determined
based on the target area and subject to various constraints. This has been
addressed in \cite{6045299} through a cell decomposition that \revised{tracks}
cell age and applies a combination of heuristics and
optimization. Also, regions have been weighted on features of interest
\cite{alamdari2014persistent} to minimize the maximum weighted latency
and to produce paths of length within a logarithmic factor of the
optimal. A related problem is to determine the minimum number of
robots. \revised{This} is approximated in \cite{asghar2019multi} within $O(log
\rho)$, $\rho$ being the ratio between maximum and minimum
latency. The problem is posed as a Minimun Cycle Cover (MCCP) to
partition the graph and, through a heuristic, to produce a set of walks
that visit the maximal number of urgent vertices.

In contrast to these methods that use age or latency as a soft
constraint and encode it in the objective function, we consider the
latency as a hard constraint. We also explicitly consider the battery
constraint in contrast to
\cite{smith2011persistent,alamdari2014persistent,asghar2019multi} or
to \cite{6045299} which consider it through reactive control policy
modification.



\section{Persistent UAV Covering}
\label{sec:method}

Persistent UAV service consists of \revised{the} continuous operation of a swarm of UAVs which are dispatched from service stations to cover a set of tasks. 
At this level, task scheduling considers UAV flight travel distance as a measurement of energy consumption, although  flight dynamics can be included in energy modeling. 
Our goal here is to minimize the number of UAVs while all requesting cells for covering tasks are visited.


\subsection{Problem Definition} 

\jml{We define the persistent covering problem as a graph covering
  problem, in which each node can be a sub-region of an area to be
  serviced. Given an undirected graph $G=\{V,E\}$, where each node
  $v\in V$ has a value $t_v$ indicating the elapsed time since the
  last service. The value $t_v$ is also known as the `age' of the node
  $v$.  Each node $v$ also has a deadline $T_v$ so at all time we need
  to ensure that $t_v<T_v$ within the planning horizon
  $H=\infty$. That is to say, every node must be serviced \textit{at
    least} every $T_v$ time units.  The weight of each edge $e \in E$
  indicates the energy needed to travel on $e$.
Such a graph $G$ can be constructed using methods such as \prm\ \cite{kavraki1996probabilistic}. 
}

%

Let each UAV $r$ have a limited available amount of battery level (i.e., a budget $b_r$ of flying time), and the battery  needs to be recharged at a charging station before the next mission.
Because the time needed to charge the batteries is almost always longer than the flying time for most consumer-level UAVs, let us assume that the time to get fully charged be $B_r > b_r$.

Let us now consider the case of a single UAV. 
The behavior of this UAV would simply be:
\begin{center} cover$\rightarrow$recharge$\rightarrow$cover$\rightarrow$recharge$\rightarrow$...., \end{center}
 and  this is repeated until all nodes are covered. It is likely that some regions may require service before this UAV is fully charged, thus,
persistent covering cannot be achieved. 
Therefore, given a persistent covering problem $G$, 
\revised{we would like to answer the following questions. First, how many UAVs are needed?. Next, can $k$ UAVs persistently cover $G$?. If so, find a schedule for these $k$ UAVs.}


\jml{
In this paper, we consider a fully predictable and simplified system: (a) All nodes have the same deadline $T$, (b) all UAVs have the same battery power $b$, (c) all UAVs need to be charged for the same amount of time $B$ regardless of the remaining power upon arriving  the charging station, 
\revised{and (d)} all UAVs start the mission from a single charging station. 
}
Although it is straightforward to relax \revised{assumptions (a) and (b), they keep our discussion} simpler. 
To this end, we answer the following question:
    What tours in a \revised{given $G$ admit the minimum number of UAVs needed to solve it?} 

Note that, for the rest of the paper, we use the term tour interchangeably with  \emph{closed walk}, 
where  a walk  in graph theory is defined as a sequence of nodes that may be repeated multiple times and a closed walk is a walk starting and ending at the same node, which in this paper is the charging station.

\begin{table*}
\centering
\caption{Symbols used in this paper. }
\label{table:symbols}
{
\revised{
 \begin{tabular}{| l | l || l | l |} 
 \hline
\textbf{Symbol}  & \textbf{Description} & \textbf{Symbol}  & \textbf{Description} \\ \hline
$G=\{V,E\}$ &  graph to be covered &  $V=\{v_i\}$  & vertices of $G$ \\ \hline
$E$ &  undirected edges of $G$ & $r$ & a single UAV    \\ \hline
$T_v$ & latency constraint of node $v$ & $t_v$ & age of node $v$ \\ \hline
$B_r$ & recharging time of $r$ & $b_r$ &  battery level of $r$ \\  \hline
$\tau$ & a closed walk (tour) of $G$  &  $P_v$ &  tours  passing through $v$   \\ \hline
 $S_i \subset \tau$ & a segment of $\tau$ & $x_i$ & decision variable of $\tau_i$ \\  \hline
\end{tabular}
}
}
\end{table*}

\subsection{An Overview of the Proposed Method}

The problem of persistent covering has been traditionally formulated as a large combinatorial  optimization problem that can be solved using  Integer  Programming (IP) solvers \cite{Song2016}. 
We aim to reduce the size of the IP by introducing a new formulation. 
The key to achieve this goal is through \revised{the extraction and filtering} of
feasible routes from the graph $G$ before the IP solver is evoked.
Under this framework, we now briefly discuss a new approach that  will be used to design multiple methods in the following section.

Let $\tau_i=\{v_0, v_1, \cdots, v_k, v_0\}$ be a tour that starts and ends at a charging station $v_0$, where $v_i \in V$. 
\revised{An example of such a tour is a solution to TSP of $G$. It is also possible to  partition $\tau_i$ into multiple segments $S_j \subset \tau_i$ and create more tours.}
We say that a tour is feasible if a UAV can complete the tour before its battery is depleted, i.e., $time(\tau_i) \le b$, and, for each node $v_i$ along the tour, the time to reach $v_i$ is less than $T$. 
To define our optimization problem, we first define the \textbf{binary decision variable} $x_i$ of the model:

\[ x_i =
  \begin{cases}
    1      & \quad \text{if } \tau_i \text{ is assigned to a UAV}\\
    0      & \quad \text{otherwise} 
  \end{cases}
\]


\revised{Now, we define the} \textbf{scheduling constraints} as follows.
%
%
Let $P_v$ be a set of tours that pass through vertex $v$ before $t_v \le T$. 
To ensure that the assigned tours cover the graph $G$, 
the following constraint is defined for each vertex $v$ of $G$:
\revised{$\sum_{ \tau_i \in P_v} x_i >0$}

Since we offload the other constraints, in particular the energy constraints, to the computation of the feasible tours, the number of constraints in this optimization problem is significantly smaller, namely $||V||$. 
The remaining step to ensure that the IP can be solved efficiently is to enforce the sparsity of the model, which depends on the length and the number of the tours. 

The \textbf{objective function} can be defined in many ways. This paper mainly focuses on minimizing the total number of UAVs needed:
\begin{equation}
\textrm{Minimize} \sum_{i}  x_i\ .
\label{eq:object}
\end{equation}
  
However, the objective function can be easily modified to minimize the total travel time in terms of distance travelled on the assigned tours:
\revised{$ \textrm{Minimize} \sum_{i}  x_i \cdot time(\tau_i).$}
Moreover, we can also minimize the overlapped coverages:
\revised{$ \textrm{Minimize} \sum_{i}  x_i \cdot card(\tau_i), $}
where $card(X)$ is the cardinality of set $X$.
 
 The IP problem formulated here is indeed a set cover problem, which admits some known strategies for increasing computational efficiency, such as for low-frequency systems in which polynomial time  approximation is possible. 
\revised{To ease our discussion, Table~\ref{table:symbols} summarizes the introduced symbols.}
 In the rest of this paper, we will focus on how the feasible tours can be determined and computed efficiently from $G$.

\section{Our Method}
\label{sec:ourmethod}

\subsection{Problem Analysis}

\subsubsection{Persistent Coverage of Small Graphs}
\jml{Let us first analyze the case of one UAV on a small graph $G$. 
We denote this UAV $r$, and
we say that $G$ is small if a TSP  $\pi$ of $G$ can be  toured by $r$ in a time at most $b$, i.e., $time(\pi) \le b$. 
We say that a node $n$ can be reached by $r$ if the path from the charging station to $n$ takes time less than $b/2$.
Given a small graph $G$ and a UAV $r$, we say that $r$ can persistently cover $G$  \textbf{if and only if}}
\begin{enumerate}
\item All nodes in $G$ are reachable in $b/2$ time. 
\item $T \ge B + b$
\end{enumerate} 

The first constraint is obvious. If the UAV cannot reach a node in $b/2$ time, it cannot be back to the charging station in $b$ time to continue its service. 
The second constraint  $T \ge B + b$ can be derived from the fact that, given a TSP tour $\pi$ of $G$, $r$ can visit all nodes in minimum time $time(\pi)$, 
namely the time needed for $r$ to finish traveling on $\pi$.
Therefore, after a node is visited, the second time that the node is visited will take $B + time(\pi)$, which must be less than $T$ to ensure continuous service of the node.
Consequently, if the UAV can visit all nodes of $G$ without charging, i..e, $time(\pi) \le b$, it is necessary and sufficient that $T \ge B +b $.

\jml{
This simple analysis leads to several additional observations. First, $time(\pi) \le b$ implies that the location of the charging station can be an arbitrary node in the graph, 
i.e. the selection of the charging station does not affect the UAV's ability to cover $G$ persistently with one UAV. Second, if the first constraint is violated, i.e., some node in $G$ is not reachable by a UAV, then 
$G$ cannot be persistently covered  even with more UAVs, unless additional charging stations can be added.
Therefore, when $k>1$ UAVs are considered, the first constraint must remain satisfied. 
}

For most consumer-level off-the-shelf UAVs, $b$ is around 15 to 25 minutes and $B$ is between one to two hours, thus $B >> b$.
Therefore, even for small graphs, the assumption that the service deadline $T \ge B + b$ is too restrictive for many real world problems. 
Naturally, we now consider the case  that $b \le T \le B+b$. This is the case that one UAV fails to cover $G$ persistently. 
Because $T \ge b$, all nodes in $G$ can be visited by the UAV $r$ before $r$ returns to the charging station and before the deadline. 
However, $B$ is too large, so we will need to send another UAV $r'$ to refresh some nodes 
before $r$ is fully charged.
Therefore, for  $k>1$ UAVs, we can send  one UAV for each $T$ time unit. To consistently cover $G$, we will need $k= \lceil(B+b)/T \rceil$.

The last case is $d \le T \le b$, where $d  < b/2$ is the travel time from the charging station to the furthest node in $G$.
In this case, some nodes might  have their timer $t$ expired before the UAV $r$ returns to the charging station. Therefore, this is the case that multiple UAVs must be sent simultaneously to cover the graph $G$.
In Section~\ref{sec:partition},  we will  discuss several strategies to handle this case.

\subsubsection{Persistent Coverage of Large Graphs}

\jml{
When a UAV takes time more than $b$  to complete a TSP  tour of $G$,  we say that $G$ is large. 
While the first condition that all nodes in $G$ should be reachable in $b/2$ time must remain true for large graphs, 
the second condition $T \ge B + b$ is insufficient for a single UAV to persistently cover large graphs since, by definition, 
the nodes cannot be toured in a time at most $b$. 
In the cases that $time(\pi) > b$ and  a single UAV, $T$ will need to be significantly larger than $B+b$.  
In the worst case, when every node in $G$ takes the UAV $b/2$ time from the charging station, $T$ will need to be at least $|V|(B+b)$ to ensure persistent coverage of $G$  by a single UAV. In essence, we must identify  subgraphs $\{G_i\}$ so that each subgraph $G_i \subset G$ is small, namely its TSP can be toured by the UAV in $b$ time, and $\bigcup_i G_i = G$. These subgraphs can then be covered by a single UAV if $T \ge card(\{G_i\})(B+b)$.
}

More realistically, if $T$ is smaller than $card(\{G_i\})(B+b)$ and we still would like to provide persistent service over $G$, 
then we need to determine what the value of $k$, the number of UAVs, must be. 
If $b \le T \le B+b$, then we will need $k=\lceil( B+b)/T\rceil$ UAVs for each subgraph and therefore $card(\{G_i\})\cdot k$ UAVs to cover $G$. 

\subsection{Creating Subgraphs}
\label{sec:partition}

When the nodes in graph $G$ have deadline $T$ that is less than the flying time $b$ or when the graph is large, 
we must partition $G$ so that multiple UAVs must  simultaneously  work on different partitions to cover $G$ persistently. 

\jml{
Let us first assume that  $two$ UAVs are available. Our goal is to create two sets $G_1 \subset G$ and $G_2 \subset G$ so that $G_1 \bigcup G_2 = G$.
It is also clear that there must be at least 2 nodes of $G$  belonging to different sets. Therefore $G_1 \setminus G_2 \not= \emptyset$ and $G_2 \setminus G_1 \not= \emptyset$.
In addition, to ensure that both $G_1$ and $G_2$ can be persistently covered, we need $time(\pi_1) < \min(T,b)$  and  $time(\pi_2) < \min(T,b)$, where $\pi_1$ and $\pi_2$ are the TSP of $G_1$ and $G_2$, respectively. Similar to the questions we asked, given a $G$, we would like to know the minimum number of UAVs needed and the subgraphs $G_k$ for each UAV.
}

In general, our graph partition problem can be formulated as the set covering problem which can be stated in a general way as follows. We are given a set of requirements or characteristics 
(say R) that must be satisfied entirely by a set of activities (say $A_x$, $A_2$,..., $A_n$) whose union equals or "cover" the entire set of requirements, and a cost associated with each activity. 
Although an activity $A_j$ may cover only a subset of R, a combination of some activities $A_j$'s may cover R. The set covering problem is to determine a combination of activities $A_j$ that can 
collectively cover all the requirements while minimizing a certain objective function. The solution of a set covering problem is defined as follows: given a set of requirements that must be fully satisfied and a set of activities each of which can satisfy some requirements and incur a certain cost, a feasible solution is defined as a select subset of activities that as a whole can satisfy all requirements. 
An optimal solution is a feasible solution with a minimal total
cost. The discussion  for the remaining of this section focus on how
to define a set of valid activities. 

%

\subsubsection{Creating Subgraphs by Segmenting a TSP of $G$}

\jml{We  find a TSP $\pi$ of $G$ then  partition $\pi$ into $k$ segments so that for each segment $S_i=\{u_0, u_1, \cdots, u_m\} \subset \pi$, the following criteria are satisfied
\begin{equation}
d(u_0)+ time(S_i) \le T , 
\label{eq:deadlineconstaint}
\end{equation}
\begin{equation}
d(u_0)+ d(u_m) + time(S_i) \le b .
\label{eq:batterycontaint}
\end{equation}
}

\begin{wrapfigure}{R}{0.2\textwidth}
\centering
\includegraphics[width=0.2\textwidth]{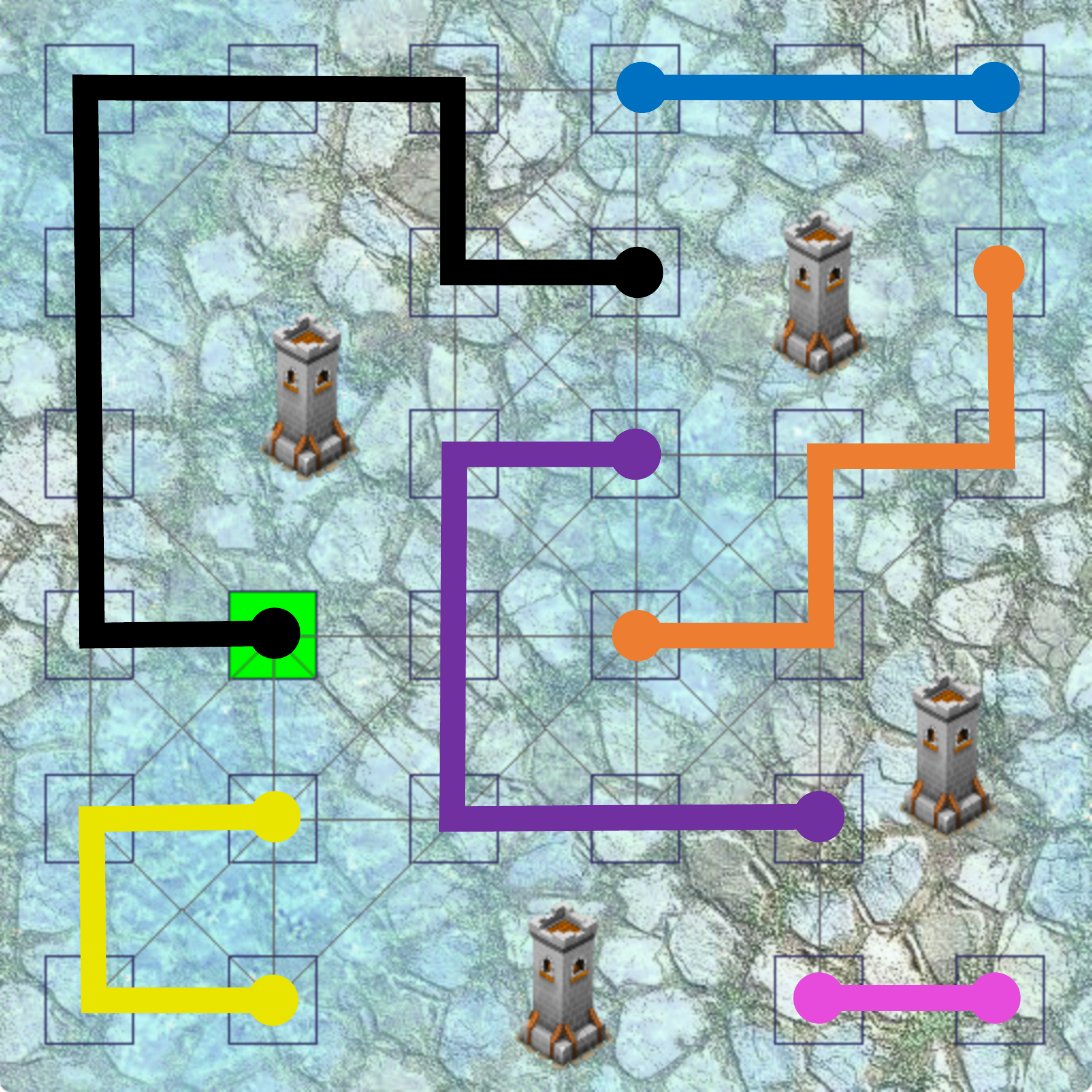}
\caption{TSP Segments.}
\label{fig:tspseg}
\end{wrapfigure}
A segment $S_i$ of $\pi$ that satisfies these properties is called a \emph{valid} segment and can be converted to valid tour $\tau_i = \{v_0 \sim u_0, u_1, \cdots, u_m \sim v_0\}$; recall that $v_0$ is the charging station node. Examples of valid segments extracted from a single TSP are shown in Fig.~\ref{fig:tspseg}. Notice that a valid segment starting from a node further way from $v_0$ must be shorter than those started closer.  
Finding these valid segments can be done greedily by starting from finding the longest $S_0$ and iterating on the remaining TSP $\pi \setminus S_0$. 
This approach is efficient but might not provide us an optimal solution  that minimizes the number of UAVs needed.  

Alternatively, we can find a longest valid segment $S_i$ that starts from each node $v_i \in \pi$. 
Segments that start at $v_i$ but shorter than $S_i$ will be superseded by $S_i$ during the optimization, thus are ignored. 
There will be $|V-1|$ of these segments therefore $|V-1|$ closed walks  $\{\tau_i\}$ from the charging station $v_0$. 
Let $\tau^*_i$ be a subset of $\tau_i$ that can be visited before time $T$. That means, $\tau^*_i$ must include the vertices $\{v_0 \sim u_0, u_1, \cdots, u_m\}$ 
and sometime a few nodes from the returning path from  $u_m$ to $v_0$. 
Then our goal is to find the smallest subset of the tours $\{\tau_i\}$  so that their corresponding $\tau^*_i$ can cover the entire $G$.
This approach guarantees the optimal solution and can be solved using 0-1 integer linear programming. 

A major drawback of this approach is that we are restricted  to a given TSP. 
Even for small environments such as those shown in Figs.~\ref{fig:env10x10} and \ref{fig:env6x6}, 
there exist many optimal TSPs. Consequently, some tours might be better than the others and yield  fewer number of UAVs needed to persistently cover the graph. 
A simple remedy is to obtain $n$ TSP paths and find the optimal subset from the $n|V-1|$ tours. However, this dramatically increases the computational cost while it is unclear what $n$ should be. In the method described in the following section, we explore a method that does not depend on the given tour.  

To ease our future discussion, we call these three TSP-based methods:
tsp-greedy, tsp-lp-1, tsp-lp-$n$, that use a greedy method, an IP
solver on $|V-1|$  longest segments originated from every node of a
TSP, and  an IP solver on  $n|V-1|$ longest segments originated from every node of $n$ TSP, respectively.

\subsubsection{Maximum Lollipop Tours of $G$}

\begin{figure}[t]
\centering
\includegraphics[width=0.475\textwidth]{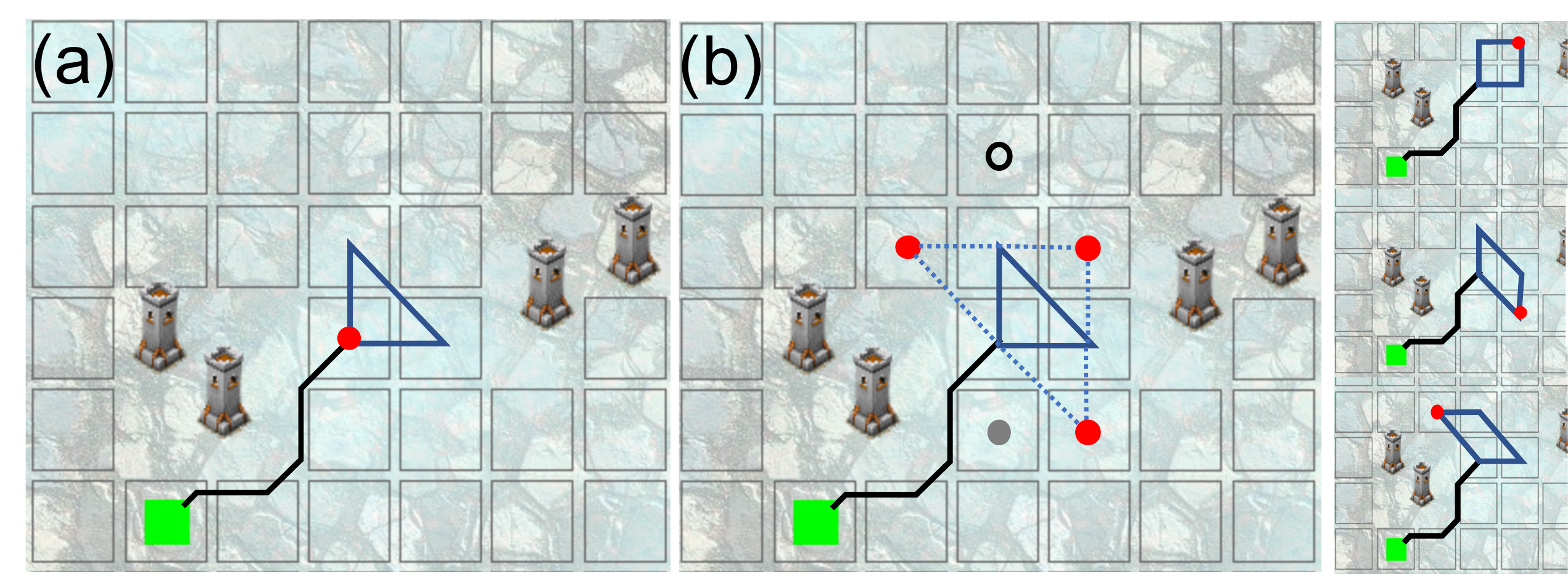}
\caption{(a) An initial lollipop  tour created from the red node. (b) This initial tour  can potentially be expanded to include one of the three incident red dots. The empty dot is not a candidate because it is only adjacent to a node in the tour and the grey dot is not a candidate because its closer to the charging station than that of its incident nodes in the tour.
Three expanded tours are shown in the right. The maximum expanded tours are shown in Fig.~\ref{fig:lollipops}.}
\label{fig:lollipop-expand}
\end{figure}

To relax from the constraint of searching on the given TSP tour(s), in this section, we discuss an approach that detects a special type of closed walk called {\em lollipop tour }
from the graph $G$. 
A lollipop tour is a closed walk  that consists of a short path from the charging station to a node $v_i$
and a cycle in a neighborhood of $v_i$   that has larger geodesic distance from the changing station than \revised{$v_i$. 
Examples }of lollipop tour can be found in Fig.~\ref{fig:lollipop-expand}.

To construct a lollipop tour for a node $v_i$, we first determine an initial cycle in a lollipop tour  at $v_i$ from a small neighborhood of $v_i$, e.g., two nodes adjacent to $v_i$ that are further away from the charging station, as shown in Fig.~\ref{fig:lollipop-expand}(a).
From the initial cycle, we iteratively  expand the cycle until the cycle has grown too large and violated the requirements.
In each iteration, we add a new node that is incident to the nodes in the cycle and is not closer to the charging station than the 
adjacent nodes in the cycle. Three of such examples are shown as the red dots in  Fig.~\ref{fig:lollipop-expand}(b). 
The lollipop tour that cannot be further expanded without violating the constraints  
is called {\em maximum lollipop tour}.  Only max lollipop tours will be considered for graph covering.

\begin{figure}[th]
\centering
\includegraphics[width=0.5\textwidth]{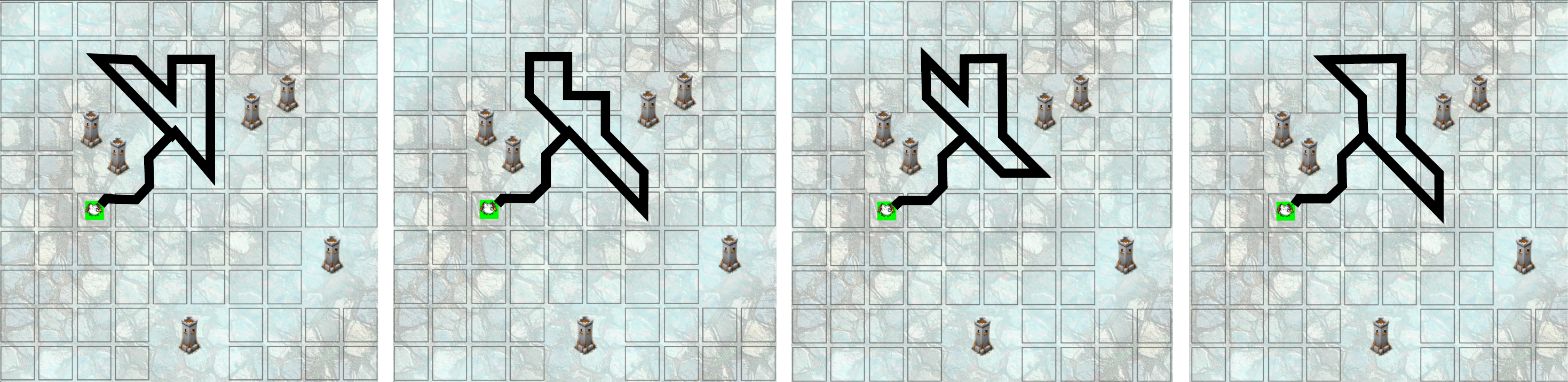}
\caption{Four maximum lollipop tours from the same node, which has 215 unique maximum lollipop tours;
$b=5K$, $B=11K$ and $T= 6K$.}
\label{fig:lollipops}
\end{figure}

As shown in Fig.~\ref{fig:lollipop-expand}(b), a cycle is likely to have  more than one way to expand. In this case, we exhaustively include all possible ways to 
expand the cycles but only keep the cycles that are unique. A pair of cycles are unique from each other if they don't share exactly the same nodes.
Uniqueness can be detected efficiently through hashing. 
In Fig.~\ref{fig:lollipops}, we show 4 maximum lollipop tours created from the example in Fig.~\ref{fig:lollipop-expand}.

Note that due to the nature of the proposed expansion process, a maximum lollipop tour might still be optimized and expanded.
If there is no new node that can be added to the cycle, we can still extract a TSP from the nodes included in the current cycle. 
The process can be repeated if the new cycle can still absorb more nodes. Otherwise the last valid cycle  is kept. 

Now that we determined all maximum lollipop tours for every node in the graph, we then find a smallest subset that covers the graph.
However, the number of unique maximum lollipop tours is exponential to the length of the cycle.  
Exhaustively enumerating all unique maximum lollipop tours is not practical for large graphs. A simple heuristic to overcome this is by gathering enough tours so that all nodes have at least one tour passing through. 

We observe that the number of maximum lollipop tours for far away
nodes is small because the length for the cycle must be small and
these tours are more critical as these are the only few tours that the
UAVs can take to reach these nodes and return safely. On the contrary,
the nodes near the charging station have many such tours but many of
them are not critical. 

Based on these facts, we sort all the nodes in descending order in terms of the distance to the charging station. All max lollipop tours are determined in this order until 
all nodes in $G$ have at least one tour passing through to ensure a solution exists from max lollipop  tours. 
Because it is possible that the entire graph is covered by max lollipop  tours without the max lollipop  tours from those nodes near the charging station, 
we add a fixed number $n$ of such tours for the nearby nodes, where $n$ is a user parameter. 
Algorithm~\ref{alg:lollipop} summarizes  this idea. Note that the subroutine $\textsc{Max\_Lollipop\_Tours}(G, v_i)$ finds all unique lollipop tours  from $v_i$ while 
$\textsc{Max\_Lollipop\_Tours}(G, v_i, n)$ terminates when $n$ tours have been identified. We use $n=10$ throughout the experiments reported in Section~\ref{sec:exp}.

\begin{algorithm}[h]
\footnotesize{
\SetAlgoLined
\KwResult{minimum number of maximum lollipop tours $\{\tau_i\}$ covering $G=\{V, E, b, B, T\}$ }
Sort $V$ in descending order in distance to $v_0$\;
Let $L=\emptyset$\;
 \For{$v_i$ in $V$}{
 
	  \eIf{$L$ does not cover $G$}
	  {
  		$L= L \bigcup \textsc{Max\_Lollipop\_Tours}(G, v_i)$ \;
	   }{
  		$L= L \bigcup \textsc{Max\_Lollipop\_Tours}(G, v_i, n) $\;
	  }
 }
$\{\tau_i\}=$ find smallest subset of $L$ that covers $G$ using IP\;
\caption{Min-Max Lollipop Tour Coverage}
\label{alg:lollipop}
}
\end{algorithm}



\subsubsection{Creating Subgraphs from shortest paths in $G$}

To compare to the proposed methods in terms of computational cost and solution optimality, 
we build a baseline method using the Dijkstra tree $D$ that encodes all shortest paths from the charging station to  the other nodes in $G$. 
From $D$, we can extract $|E| - |V-1|$ loops, namely a loop for 
each edge in $G$ that does not belong to $D$. We then keep the valid loops that satisfy the constraints (Eqs.~\ref{eq:deadlineconstaint} and ~\ref{eq:batterycontaint}). 
Note that it is also possible to have a trajectory for an UAV that has identical departure and return paths.  There will be $|V-1|$ of such paths, thus the total number of closed walks is $|E|$. We then find an optimal subset sum from the valid loops of these $|E|$ closed walks using an IP solver.

%
%
%
%
\section{Experiments}
\label{sec:exp}

\begin{figure}[th]
\centering
\includegraphics[width=0.5\textwidth]{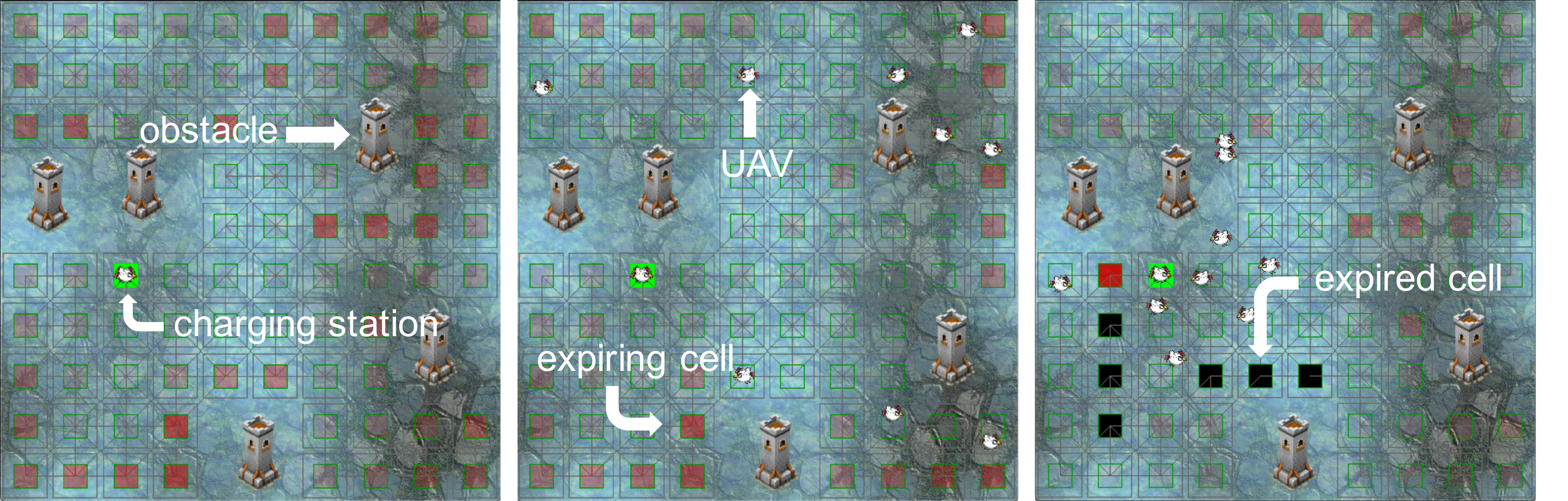}
\caption{Experiment setup for persistent covering.}
\label{fig:env10x10}
\end{figure}

\revised{Minimizing the number of UAVs needed for covering a graph persistently has strong economic incentives. 
The maintenance cost and coordination complexity can be dramatically reduced even if the scheduler can reduce a single UAV. 
Therefore, in our experiments, we compare the number of UAVs produced by the proposed methods using various graph sizes and types.}
We implement all methods described in Section~\ref{sec:method} in C++ and use Concorde as the  TSP solver and GLPK as the integer programming solver.  
The running times reported in this section are collected from a laptop computer. 
\revised{All experiments reported in this section are setup as illustrated in Figs.~\ref{fig:uas} and \ref{fig:env10x10}. }
We refer the reader  to the \revised{accompanying} video for better visualization.



\begin{figure}[t]
\centering
\includegraphics[width=0.4\textwidth]{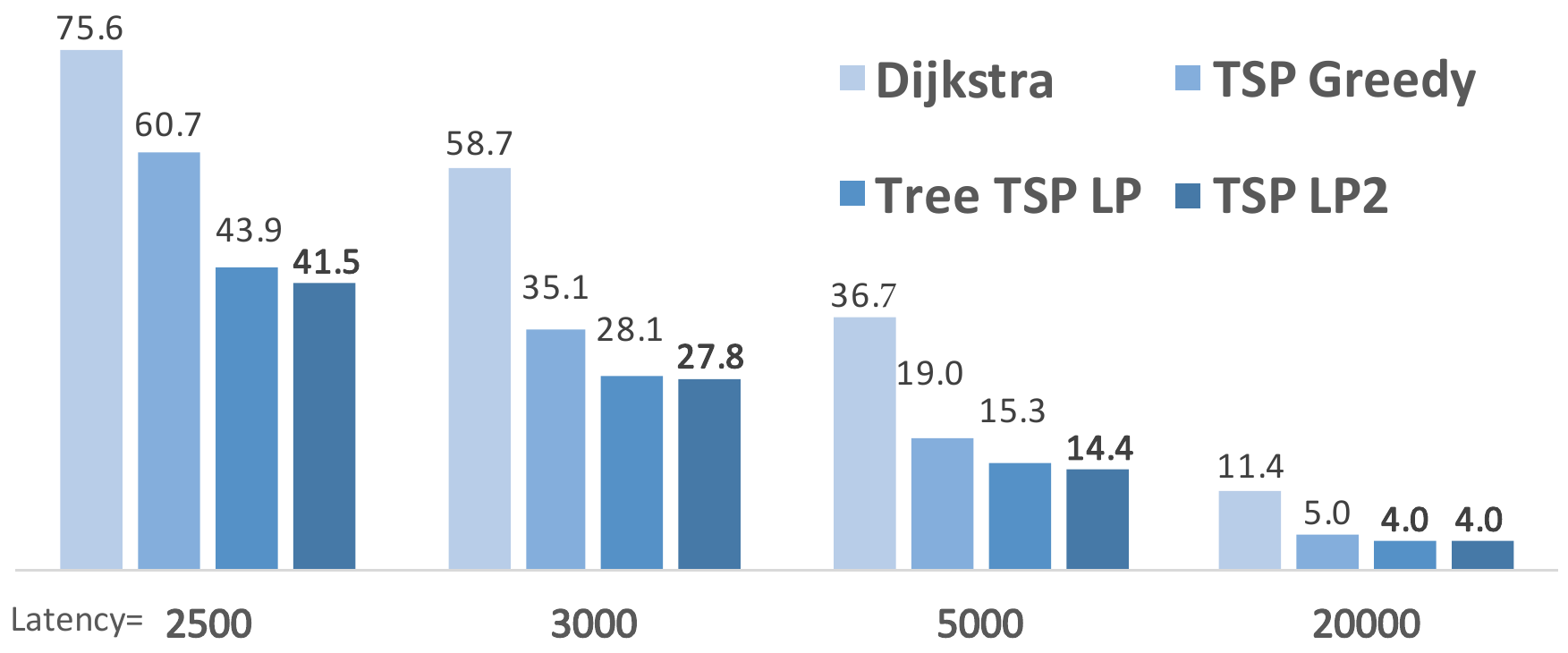}
\caption{Number of UAVs needed to cover a 6x6 environment with various latencies. Data reported as an average of 10 trials. In all cases, $b= 5,000$s and $B=11,000$s.}
\label{fig:6x6}
\end{figure}

\begin{figure}[t]
\centering
\includegraphics[width=0.45\textwidth]{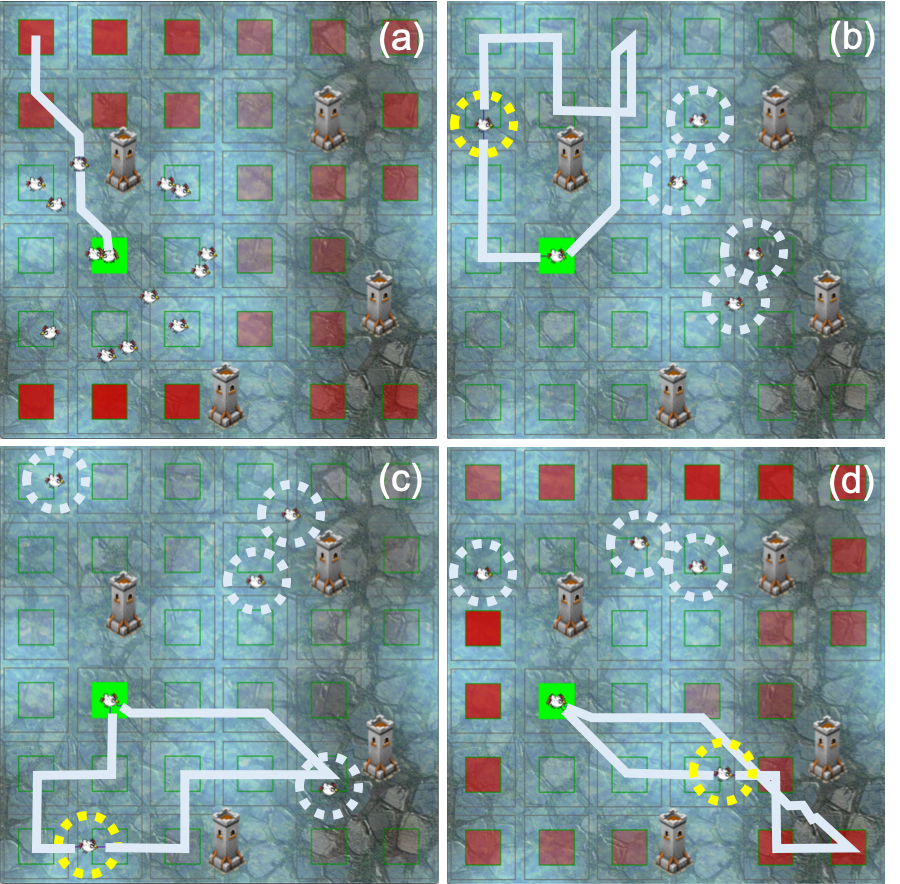}
\caption{A $6\times 6$ environment used for persistent covering. (a) All shortest paths from the charging station.  39 UAVs are needed. 
(b) TSP greedy  with 10 trials; 15 UAVs are needed. (c) TSP LP with 10 trials; 15 UAVs are needed. (d) TSP LP will aggregated 10 trials; 12 UAVs are needed. }
\label{fig:env6x6}
\end{figure}

We \revised{first} test our implementations on \revised{the environments shown in Figs.~\ref{fig:env10x10} and ~\ref{fig:env6x6} with battery time = 5,000 seconds and charging time = 11,000 seconds}. 
The battery time is roughly a half of the charging time. 
Fig.~\ref{fig:6x6} presents the results from the $6\times6$ field illustrated in Fig.~\ref{fig:env6x6} by varying the latency constraints from 2500 seconds to 20,000 seconds. 
It's obvious that when the latency constraint is larger fewer UAVs are needed. The baseline method that extracts tours using Dijkstra tree requires many more UAVs than the TSP-based methods in all cases. For all TSP-based methods, we use Concorde to extract 20 unique TSP tours from the environments. 

\begin{figure}[ht]
\centering
\includegraphics[width=0.45\textwidth]{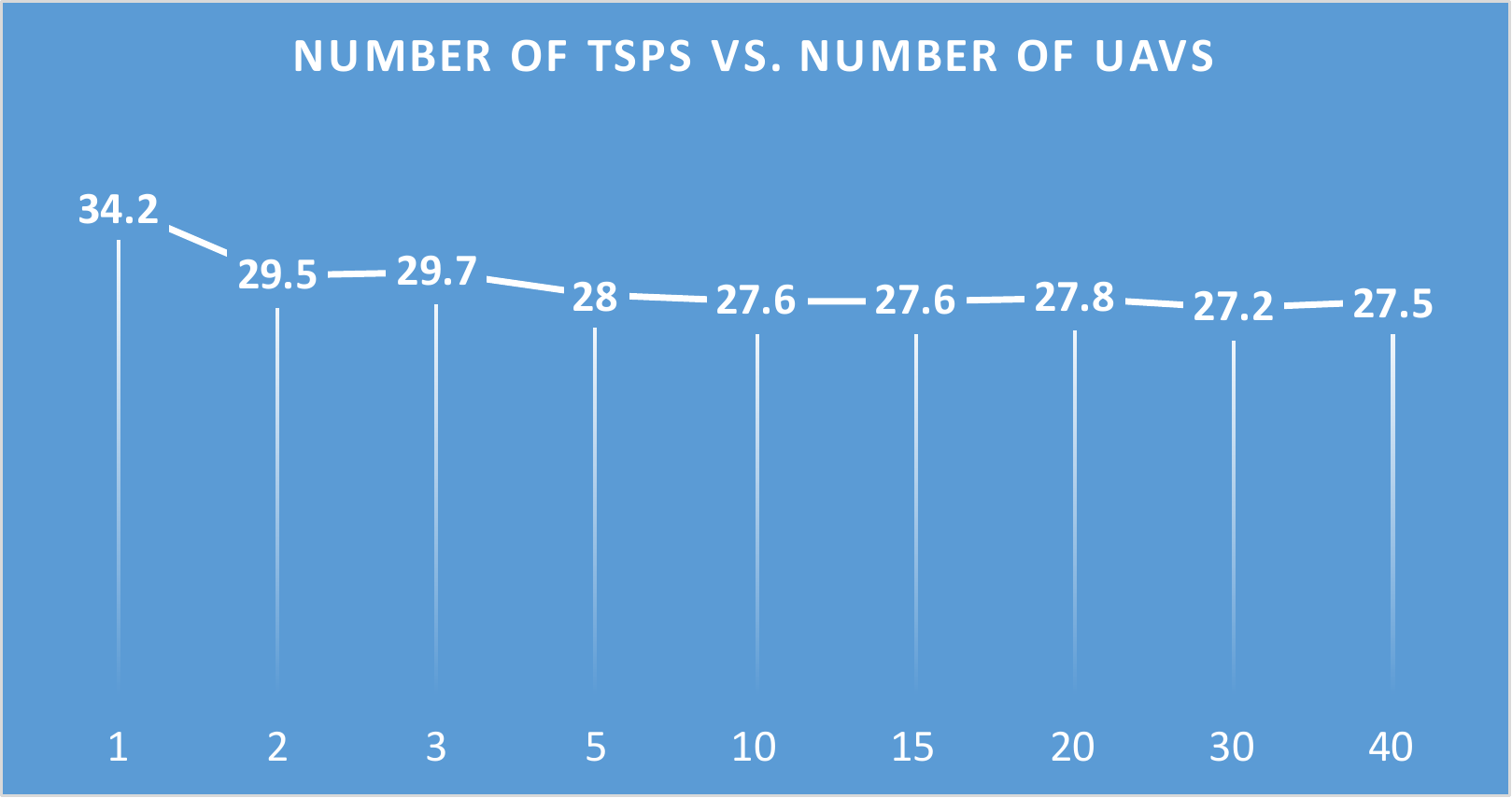}
\caption{Number of TSP tours considered ($x$-axis) vs. number of UAVs needed ($y$-axis).}
\label{fig:tspneeded}
\end{figure}

Fig.~\ref{fig:tspneeded} justifies the choice of 20 by showing that the number of UAVs required for persistent covering stabilizes when more than 10 TSP tours are used.
As we can observe from the data, when we segment each of TSP paths greedily,  the minimum number of UAVs needed  is 
noticeably larger than the other two methods and shows that optimization is indeed necessary. 
When we exhaustively segment each individual TSP and solve the set cover problem using integer programming (i.e., TSP  LP), 
the minimum number of UAVs needed  is slightly larger than solving the set cover problem using all 20 tours (i.e., TSP  LP2). 
The running times of Dijkstra tree, TSP greedy, TSP LP, TSP  LP2 are 0.25 secs, 0.49 secs, 0.54  secs, and 0.55 secs, respectively.

\revised{To test the proposed methods on the larger graph,} we repeat the same experiment \revised{using the $10 \times 10$ field in Fig.~\ref{fig:env10x10}}. 
The results are reported in Table~\ref{table:10x10}.
We see a similar pattern that the Dijkstra tree method, with the lowest running time, requires significantly more UAVs to cover the field persistently. 
Unlike the 6x6 example, we found that TSP LP2 performs on par with TSP LP when latency is larger or equal to the battery time but noticeably outperforms
TSP LP when the latency is smaller than the battery time.  The best of all methods that requires the minimum UAVs is the method, called Hybrid, 
that combines TPS tours and maximum lollipop tours. As we have shown earlier that increasing the number of TSP tours beyond 10 does not reduce the number of UAVs,
however adding lollipop tours consistently increases the diversity of
the population, thus resulting in a better solution. 

\begin{figure}[h]
\centering
\includegraphics[width=0.45\textwidth]{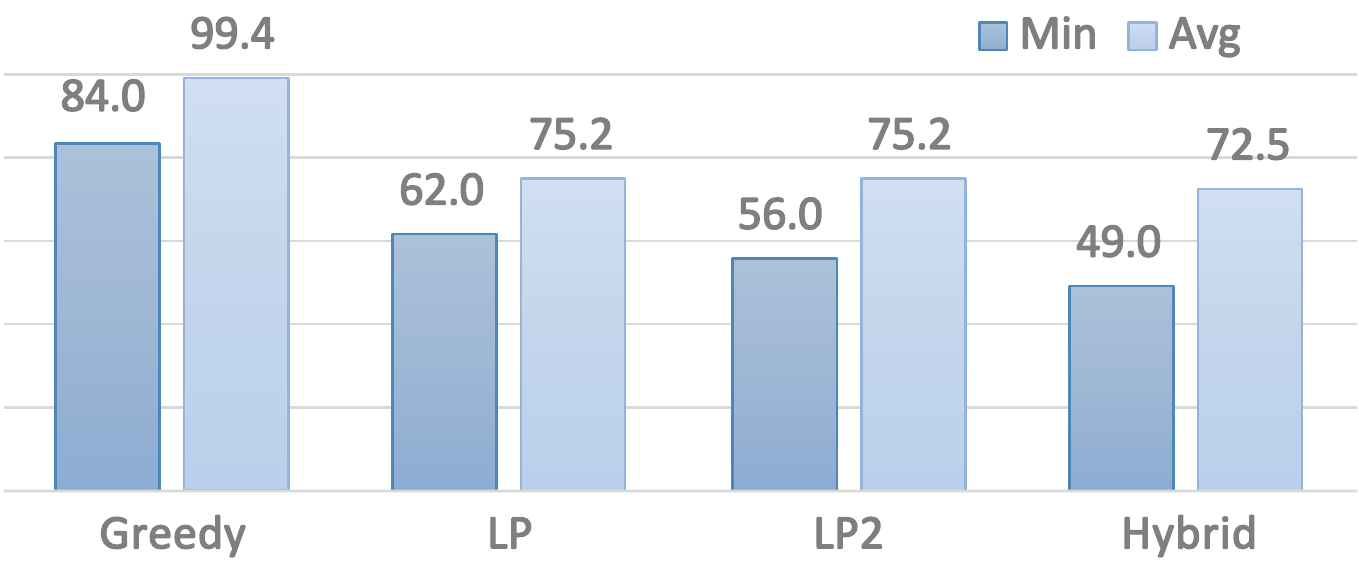}
\includegraphics[width=0.45\textwidth]{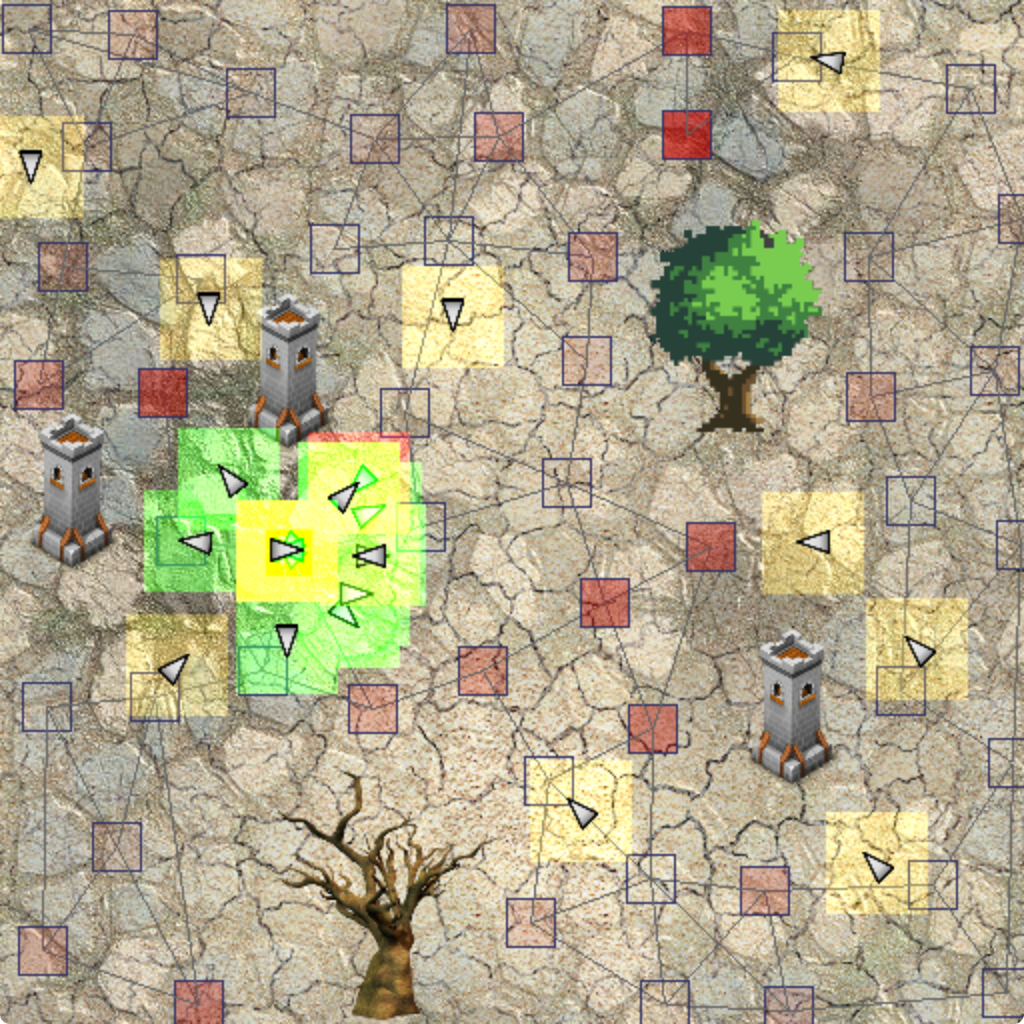}
\caption{A 10x10 environment with a  \prm\ \cite{kavraki1996probabilistic} graph .}
\label{fig:10x10env}
\end{figure}
The benefits of the hybrid method can be further seen when the structure of the graph is less regular as in roadmaps obtained for probabilistic methods. 
In Fig.~\ref{fig:10x10env}, we see that, again, TSP LP2 performs on
par with TSP LP. While the hybrid method  requires a significantly lower number of UAVs. 
Finally, persistent covering a $20 \times 20$ field is demonstrated in the accompanied video. The hybrid method takes  around 13 minutes to determine that 75 UAVs are needed.


%
%

\section{Conclusion}

This paper presented the first method that can determine the minimum number of UAVs needed for persistently covering 
 under the constraints of flight endurance, battery charging time and  time between revisits.
The proposed framework identifies segments of TSP and lollipop tours with respect to the energy constraints. A subset of these tours then forms the minimum cover of the graph that satisfies the latency constraint. The experiment results show that the proposed framework produces schedules that require significantly  fewer UAVs than the baseline methods. 
\revised{Field work will be conducted in the near future to deploy and evaluate our schedules on Unmanned Aircraft Systems}.
It should be noted that, even though all examples in the paper are solved in less than a few minutes,  
the time complexity of the presented method is still naturally exponential to the size of the graph. \revised{Our future work will relax assumptions made in this paper to include multiple or mobile charging stations, heterogenous UAVs, and more realistic energy and kinodynamic model of the UAVs. }

\begin{table}
\centering
\caption{Experiments on a 10x10 environment with varying latency. }
\label{table:10x10}
{
 \begin{tabular}{||l | l||r r r r r||} 
 \hline
               & & Dijkstra  &TSP  & TSP  & TSP    & Hybrid \\ 
Latency   & & Tree & Greedy & LP & LP2 & \\[0.5ex] 
 \hline\hline
               &  $N$ &  32  & 10.00  & 7.60 & 7.50 & \textbf{7.00}\\
$20000$ & $K$ & 32 & 10.00 &  7.60 & 7.50 & \textbf{7.00}\\ 
             & time  &  \textbf{1.31} & {1.68} & 1.75 &4.38 & {2.59}\\
 \hline
             & $N$ &  97  & 35.00  & 29.80 & 28.30 &\textbf{26.67}\\
$5000$ & $K$ &30 &  9.00 &  7.70  &  7.50 &  \textbf{7.00}\\
             & time  & \textbf{1.21}   & {1.67} & 1.83 &1.81 & {2.45}\\
 \hline
             & $N$  &167 & 66.80 & 55.80 & {50.20} & \textbf{48.67} \\
$3000$ & $K$ & 32 & 9.90 & 10.00 & {8.90} & \textbf{8.78}\\
             & time  &  \textbf{1.23} & 1.57& 1.84 &  1.94 & {1.97}\\
 \hline
             & $N$  &177 &  114.90 & 88.30 & {76.50} & \textbf{76.33} \\
$2500$ & $K$ & 29 & 17.90 & 1.98 & {11.80} & \textbf{8.78} \\
             & time  &  \textbf{1.16} & 1.61 & 1.98 & 2.06 & {2.22}\\
 \hline
\end{tabular}
}

{\footnotesize  Note: In all cases, $b= 5000$ and $B=11000$. $N$ is the number of UAVs and $K$ is the number of tours needed for persistent covering, and ``time" is running time in seconds.
All data are the average of 10 runs.} 
\vspace{-0.8cm}
\end{table}

\out{
\section{Discussion and Additional Optimization}

So far, we have focused on finding a strategy that first finds a tour that ensures a UAV can finish and then determines the number of UAVs needed for this tour to satisfy the latency constraint.  This means that the number of UAVs is tightly coupled with the tour and, consequently, some nodes in the graph might be visited more frequently than needed.
In this section, we discuss another approach that consider finer granularity in scheduling the persistent tours.

Let us now consider a node $n$ of the graph $G$ and let $L=\{\ell_i\}$ be a set of tours (e.g., max lollipop tours) passing through $n$ including those that may arrive $n$ after $T$.
It is important to know that the tour $\ell_i$ is slightly different, i.e., each tour is associated with a single UAV only and has associated arrival time for
each node it visits. We call such your a timed tour. Therefore, each tour $\tau$ can result in $\lceil(B+b)/T \rceil$ timed tours $\ell_i$ with each $\ell_i$'s departure time offset by $T$. 

Our goal here is to determine all subsets of $L$ each of which  permits  persistent coverage of $n$.
Let $K=\{\kappa_0 \cdots \kappa_k\} \subset L$  be such a subset which is organized as a  list of tours ordered by their arrival times.
We further let $A=\{a_0 \cdots a_k\}$ be the corresponding arrival times  of $\{\kappa_i\}$  at $n$.
\jml{For this  method to work, we also need to make sure that all tours in $K$ must have the same tour time, i.e., $ t_K =  \max \left( \mathrm{time}(\kappa_i) \right) \le b$.}
To ensure that $K$ provides  persistent coverage for $n$, we must ensure that 
\begin{multline}
a_{i+1}+\mathrm{time}(\kappa_{i+1}) -a_i-\mathrm{time}(\kappa_i)  \\ 
= a_{i+1} - a_i \le T, \mathrm{ for\ all\ } 0 \le i \le k-1 \ ,
\end{multline}

and
\begin{equation}
(a_0 + \mathrm{time}(\kappa_0) + B)-a_k \\
= a_0 + t_K+B - a_k \le T \ . 
\end{equation}

Let $\{K_j\}$ be a set of all possible $K$s from $n$ and  $y_j$ be the decision variable for $K_j$. Then, we obtain the following additional constraints 
$$
\mathrm{card}(K_j) y_j \le \sum_i x_i \ ,
$$
where $x_i$ is a decision variable indicating if a tour $\kappa_i$ in $K_j$ is selected. This constraint essentially states that if $y_j$ is 1, then all tours in $K_j$ must be selected.
Otherwise, we don't care if the tours in $K_j$ are selected or not. 
To ensure that the node $n$ is covered by at least one of such $K$s, we constraint the  system: 
$$
\sum y_j \ge 1 \ .
$$

\todo{Need to describe how $\{K_j\}$ is computed}
The IP solver simply takes these additional constraints for all nodes in $G$ with the same objective function defined in Eq.~\ref{eq:object}.
The only difference is that each tour is only associated with UAVs.

}

\bibliographystyle{IEEEtran}
\bibliography{georobotics,geom,energy-awareness}

\end{document}